# Neural inhibition during speech planning contributes to contrastive hyperarticulation

Michael C. Stern[a,b] & Jason A. Shaw[a]

[a]Department of Linguistics, Yale University, 370 Temple St, New Haven, CT, USA

[b]Corresponding author: michael.stern@yale.edu

**Abstract**

Previous work has demonstrated that words are hyperarticulated on dimensions of speech that differentiate them from a minimal pair competitor. This phenomenon has been termed contrastive hyperarticulation (CH). We present a dynamic neural field (DNF) model of voice onset time (VOT) planning that derives CH from an inhibitory influence of the minimal pair competitor during planning. We test some predictions of the model with a novel experiment investigating CH of voiceless stop consonant VOT in pseudowords. The results demonstrate a CH effect in pseudowords, consistent with a basis for the effect in the real-time planning and production of speech. The scope and magnitude of CH in pseudowords was reduced compared to CH in real words, consistent with a role for interactive activation between lexical and phonological levels of planning. We discuss the potential of our model to unify an apparently disparate set of phenomena, from CH to phonological neighborhood effects to phonetic trace effects in speech errors.

# 1 Introduction

*1.1 Contrastive hyperarticulation*

Previous work has shown that the pronunciation of a word is gradiently influenced by its relationships with other similar words in the lexicon. For example, phonological neighborhood density (Luce, 1986; Vitevitch & Luce, 2016)—defined broadly as the number of similar-sounding words in the lexicon (Vitevitch & Luce, 2016)—has been shown to affect the pronunciation of both vowels (Gahl et al., 2012; Munson, 2007; Munson & Solomon, 2004, 2016; Scarborough, 2010, 2012, 2013; Wright, 2004) and consonants (Fox et al., 2015; Fricke et al., 2016). The number of similar-sounding words in the lexicon seems to influence pronunciation. Phonological neighborhood effects on pronunciation can be characterized as *global*, since they are not targeted towards specific lexical competitors. However, a more *local* effect is caused by the presence of a minimal pair competitor. We consider a minimal pair competitor to be a word that differs from the target word in a single dimension of speech or "phonological feature". The main phonological feature of interest in this paper is "voicing", which differentiates, e.g., the initial consonant of PET[1] from that of minimal competitor BET. Consonant voicing corresponds articulatorily to the temporal coordination between oral and laryngeal gestures; voicing can be measured acoustically as the duration of the interval between the "burst" caused by the sudden release of an oral occlusion (e.g., of the lips for /p/[2]) and the onset of periodic vocal fold vibration associated with the following vowel. This duration is termed "voice onset time" (VOT: Abramson & Whalen, 2017; Lisker & Abramson, 1964). In conversational speech, "voiceless" consonants like /p, t, k/ are defined by a long VOT (about 40-80 ms), while "voiced" consonants like /b, d, g/ are defined by a short VOT (about 10-30 ms) (e.g., Chodroff & Wilson, 2017).

Change in the pronunciation of a word that specifically differentiates it from a minimal pair competitor—termed "contrastive hyperarticulation" (Wedel et al., 2018)—has been observed in several specific dimensions of speech, including vowel formants (Clopper & Tamati, 2014; Wedel et al., 2018), vowel duration (Goldrick et al., 2013; Schertz, 2013; Seyfarth et al., 2016), and voicing of voiceless consonants (Baese-Berk & Goldrick, 2009; Buz et al., 2016; Goldrick et al., 2013; Nelson & Wedel, 2017; Schertz, 2013; Wedel et al., 2018) and voiced consonants (Nelson & Wedel, 2017; Schertz, 2013; Seyfarth et al., 2016; Wedel et al., 2018). CH is magnified when the minimal pair is contextually salient (Baese-Berk & Goldrick, 2009; Kirov & Wilson, 2012; Seyfarth et al., 2016) and after the listener indicates that they heard the minimal pair competitor rather than the target word (Buz et al., 2016; Schertz, 2013). CH may underlie the influence of functional load—operationalized as the number of lexical items distinguished by a phonological feature (Hockett, 1967)—on articulation (Hall et al., 2017). This influence has been argued to maintain the distinctiveness of words over time (Wedel, 2012; Wedel et al., 2013; Winter & Wedel, 2016). While some have argued that global neighborhood effects are in fact reducible to local CH effects (Buz & Jaeger, 2016; Nelson & Wedel, 2017; Wedel et al., 2018) or vice versa (Fox et al., 2015; Fricke et al., 2016), others have argued that the two types of effects arise from (at least partially) independent sources (Clopper & Tamati, 2014). The focus of the present study is CH, rather than general neighborhood effects; however, the results have implications for understanding the relationship between these types of effects, a point we return to in the discussion section.

---

[1] All caps indicates a lexical representation.
[2] Slashes indicate a phonological representation.

Baese-Berk & Goldrick (2009:Experiment 2) offers a representative example of an experiment demonstrating CH. During each trial in this experiment, participants were shown a screen with three words, one of which (the target) was highlighted. The participant was instructed to read the target word to a "listener" (a lab confederate) who was viewing a different screen, identical to the participant's but without the target word highlighted. The participant's goal was to get the listener to select the correct target word. All target words were real words beginning with voiceless stop consonants (e.g., /p, t, k/), but they were divided into three conditions. Words in the "no competitor" condition had no minimal pair competitor differing in initial consonant voicing, e.g. PIPE (*BIPE)[3], and the two distractor words were unrelated to the target. Words in the "no context" condition had a minimal pair competitor, e.g. PAD (BAD), but this minimal pair was not presented as a distractor; again, the two distractors were unrelated to the target. In the "context" condition, the target's minimal pair competitor was presented as an on-screen distractor. Figure 1 displays the results. The VOT of words in the "no context" condition was hyperarticulated compared to those in the "no competitor" condition by an average of about 5 ms, phonetically differentiating the voiceless target from the voiced minimal pair competitor. Moreover, words in the "context" condition were hyperarticulated compared to those in the "no competitor" condition by about 10 ms, demonstrating an additional effect of the contextual salience of the minimal pair competitor.

---

[3] The asterisk indicates absence from the lexicon.

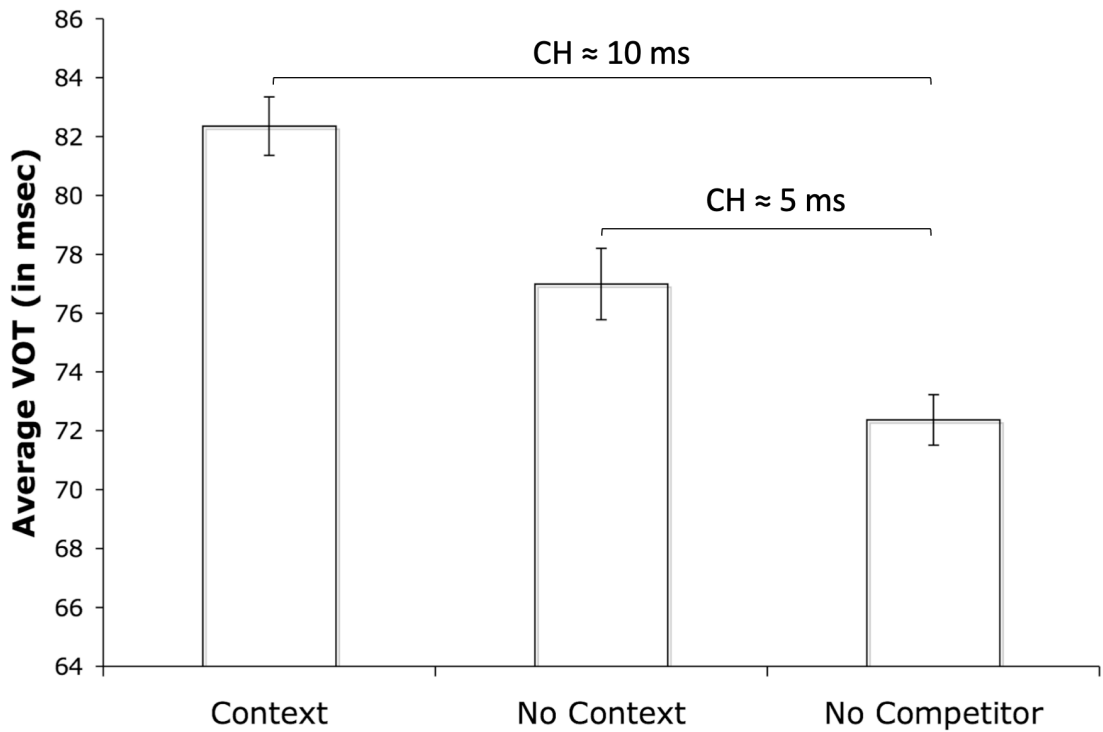

**Figure 1.** Results of Baese-Berk & Goldrick (2009:Experiment 2).

*1.2 Possible sources of contrastive hyperarticulation*

A number of hypotheses have been proposed to explain CH. On one hand, CH may arise from relatively slow processes occurring over the lifetimes of language users. In episodic memory models of linguistic knowledge (Goldinger, 1998; Pierrehumbert, 2001, 2002), lexical representations are phonetically detailed and continually updated after each production or perception of the lexical item. Crucially, in many implementations of such models, only successfully categorized exemplars influence the overall lexical representation (Hay et al., 2015; Wedel, 2006). *Hypo*articulated productions of words with minimal pair competitors are more likely to be miscategorized than those of words without minimal pair competitors. Thus, over time, phonetic representations of words with minimal pair competitors will tend to become more peripheral, i.e., more differentiated from their minimal pair competitor. Episodic memory accounts of CH thus provide a possible mechanism underlying the observation that phonological category oppositions which carry a greater functional load are less likely to merge over time (Wedel et al., 2013). However, this mechanism alone cannot account for the effect of the contextual salience of the minimal pair competitor, i.e. the difference between the "context" and the "no context" conditions.

Thus, there is likely a role for relatively fast processes unfolding on millisecond timescales in the planning and production of speech. For instance, CH has been attributed to real-time *listener accommodation* on the part of the speaker (Munson & Solomon, 2004; Wright, 2004). According to this account, speakers are sensitive to the perceptual needs of the listener, and actively adjust their pronunciation in an attempt to maximize the likelihood that they will be accurately perceived by the listener (Lindblom, 1990). More broadly, this would constitute a specific case of audience design, whereby speakers take into account the likely perceptual experience of the listener (Arnold et al., 2012; Bell, 1984; Galati & Brennan, 2010). This theoretical position has been deployed to explain pronunciation variation, as in CH, as well as other language behaviors, including word choice, lexico-syntactic patterns, and semantic-pragmatic reasoning (for a review see Jaeger & Buz, 2017). With regard to CH, words with minimal pair competitors are more likely to be misperceived than words without minimal pair competitors (Vitevitch & Luce, 1998), so speakers might hyperarticulate dimensions of pronunciation that differentiate words from their minimal pair competitors, to accommodate the listener. Hyperarticulation would be further magnified when the minimal pair competitor is more salient in the context and therefore more confusable with the target. This account is supported by the finding that speakers actively adjust their pronunciations in response to an explicit misunderstanding on the part of the listener (Buz et al., 2016; Schertz, 2013).

CH has also been attributed to real-time speaker-internal processes, independent of the listener. This kind of hypothesis is based on "interactive activation" models of speech planning in which activation of one representational unit (e.g., lexical item or phoneme) causes partial activation of related representational units (Dell, 1986; Dell et al., 1999, 2021). For instance, consider the planning of the word PUN, schematized in an interactive activation framework in Figure 2. First, intention to produce PUN increases activation of its lexical representation. Via connections between lexical and phonological levels of planning, activation of the lexical representation PUN increases activation of its constituent phonological representations /p/, /ʌ/, and /n/. These lexical-phonological connections are bidirectional, so the active phonological representations send activation back to the lexical level, forming an excitatory feedback loop. Importantly, at this stage, lexical representations which *overlap* in their phonological representation with the target word also receive some activation via phonological-lexical feedback. The minimal pair BUN, differing from the target PUN in only one phonological feature (voicing of the initial consonant), receives a non-negligible amount of activation at this stage. Because of the bidirectional nature of interactive activation, the competitor lexical representation BUN subsequently sends some activation to the competitor phoneme /b/, which is not part of the phonological representation of the target word PUN. It has been proposed that partial activation of the competitor phoneme (in this case, /b/) drives hyperarticulation of the target phoneme (in this case, /p/) (Baese-Berk & Goldrick, 2009). According to this hypothesis, the different magnitudes of CH (between the "context" and "no context" conditions) can be derived from different magnitudes of activation of the competitor phoneme. In the "context" condition, the competitor receives activation from the visual context, in addition to the activation it receives from phonological-lexical feedback, driving a larger CH effect. However, the actual mechanism by which competitor activation in planning leads to CH in articulation has not been specified.

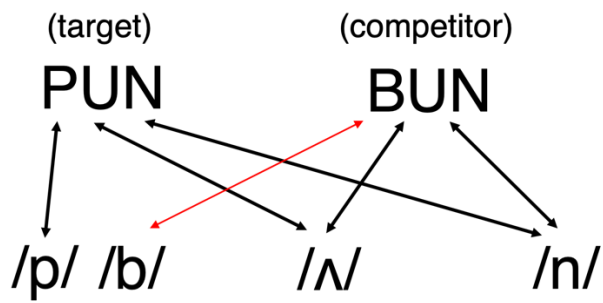

**Figure 2.** During planning of PUN, interactive activation (double arrows) causes partial activation of the competitor phoneme /b/ (red arrow).

The notion that multiple active phonological representations can simultaneously influence articulatory movement is consistent with "cascading activation" models of speech planning (Alderete et al., 2021; Goldrick & Blumstein, 2006; Goldrick & Chu, 2014). Such models have been used to explain the phonetic "trace" effect observed in speech errors, whereby errorful productions exhibit a "trace" of the intended phoneme (Goldrick et al., 2016). For instance, if BUN were produced errorfully, instead of the intended PUN, the initial /b/ would tend to have a *longer* VOT, exhibiting a trace of the intended /p/, compared to a non-errorful production of BUN. Applied to CH, this mechanism appears to make an incorrect prediction: partial activation of the competitor phoneme should cause the target word to be *more similar* to the minimal pair, i.e., hypo-, not hyper-, articulated. To address this issue, it has been proposed that the activation dynamics of competitors differ between errorful and non-errorful speech: when an error is produced, the intended target is still quite active, deriving a trace effect; in non-errorful speech, the target word dominates planning, and competitors do not receive enough activation to influence pronunciation (e.g., Baese-Berk & Goldrick, 2009:549). CH in non-errorful speech is proposed to derive from a mechanism distinct from the one deriving trace effects in errors. However, such a proposal is only necessary if cascading activation between levels of planning is assumed to be *excitatory*, pulling the pronunciation towards that of active representation(s). If a minimal pair competitor can exert an *inhibitory* influence on pronunciation planning processes, then partial activation of the minimal pair could drive *dissimilation* rather than assimilation. In this way, both trace effects and CH could derive from partial activation of competitors, the difference residing solely in the *polarity* of the influence of the competitor: excitatory (trace effect) or inhibitory (CH). Inhibitory mechanisms in speech planning have previously been proposed to account for dissimilation between simultaneously planned English vowels (Tilsen, 2007, 2009, 2013) and Mandarin tones (Tilsen, 2013). In the present study, we implement the hypothesis that inhibitory influence from a minimal pair competitor on pronunciation planning contributes to CH, using a neural dynamic model of speech planning based on Dynamic Field Theory (DFT: Schöner et al., 2016). Then, we test the predictions of the model with a novel experiment.

The remainder of the paper is structured as follows. In Section 2, we present a neural dynamic model of VOT planning, and demonstrate using simulations how the model generates CH from inhibitory influence from a minimal pair competitor. In Sections 3 and 4, we present and discuss the results of a speech production experiment designed to test predictions of the neural model. Section 5 is a general discussion, integrating the modeling and experimental results and suggesting directions for future work. Section 6 concludes.

## 2 Dynamic neural field model of VOT planning

In DFT, features relevant to perception, behavior, and cognition are modeled as continuous parameters represented by populations of neurons. The distribution of activation (spike rate) across the neurons in a population is modeled as a *dynamic neural field* (DNF). DNFs evolve over time under the influence of input (e.g., from sensory surfaces), lateral interactions, and noise. An important characteristic of DNFs is their ability to resolve competition. A single DNF can be simultaneously influenced by multiple inputs, and the dynamics of the field specify how, under the right conditions, these inputs will resolve to a single output, sometimes reflecting characteristics of both inputs. This mechanism has been shown to derive the influence of perceptual input on the initiation timing of speech movements (Roon & Gafos, 2016), the influence of distractors on the location of reaching movements (Erlhagen & Schöner, 2002), and the influence of phonological competitors on speech targets in errors (Stern et al., 2022).

In many DNF implementations, inputs to the field are strictly excitatory, increasing activation in a region of the field until a stable peak of activation forms in that region corresponding to, e.g., a particular percept or movement plan. However, it has been proposed that regions of DNFs can also be *selectively inhibited* in order to *prevent* formation of a stable peak in that region. This mechanism has been hypothesized to regulate attention in the presence of multiple salient percepts (Houghton & Tipper, 1994). Models of selective inhibition have been shown to derive the "negative priming" (Tipper, 1985) and "inhibition of return" (Posner & Cohen, 1984) effects observed in response-distractor tasks. Of particular relevance for CH, these models also exhibit *dissimilation* of movement targets from distractors in manual reaching and eye saccades (Tipper et al., 2000), as well as English vowel (Tilsen, 2007, 2009) and Mandarin tone production (Tilsen, 2013). In order to derive dissimilation from selective inhibition, it is assumed that different feature representations (modeled as DNF input distributions) *overlap* to some degree in feature space. Thus, selective inhibition of, e.g., a voiced consonant category (distributed over the lower values in a VOT planning field) causes partial inhibition of the voiceless consonant category. In particular, those neurons which are sensitive to the lowest VOT values in the voiceless distribution will be inhibited. In this way, the VOT target of a voiceless production planned during selective inhibition of the voiced category will tend to be *higher*, i.e. more hyperarticulated, compared to a production planned during no selective inhibition. Selective inhibition thus offers a mechanism by which influence of a minimal pair competitor on planning can lead to CH. In particular, inhibitory projection from a voiced competitor to a VOT planning field during production of a voiceless target is expected to cause increased VOT, i.e. CH. We demonstrate this in the following subsections.

*2.1 Data availability*

The modeling reported below was done using the MATLAB-based software COSIVINA (Schneegans, 2021). Scripts for simulating the results below are available on OSF at https://osf.io/hz8fp/.

*2.2 Model structure*

The structure of the DNF model of VOT planning presented here is mostly identical to that described in (Stern et al., 2022). The model is summarized in Eq. 1:

$$\tau \dot{u}(x,t) = -u(x,t) + h + s_{target}(x,t) + s_{mp}(x,t) + \int k(x-x')g(u(x',t))dx' + q\xi(x,t) \quad (1)$$

The key component of the model is the activation field $u$ defined over the VOT dimension $x$ at each moment in time $t$. We set the range of $x$ to 200, representing a range of possible VOT targets (in ms). The rate of change of activation $\dot{u}(x,t)$ is inversely related to current activation $u(x,t)$, so Eq. 1 represents a dynamical system with a point attractor at $h + s_{target}(x,t) + s_{mp}(x,t) + \int k(x-x')g(u(x',t))dx' + q\xi(x,t)$. $\tau$ is a time constant, with higher values corresponding to slower rates of field evolution. The resting level $h$ is assumed to be below zero for all field locations (neurons), by convention at –5. Each field input $s_{target}(x,t)$ and $s_{mp}(x,t)$ is represented as a separate Gaussian distribution of the form

$$s(x,t) = a \exp\left[-\frac{(x-p)^2}{2w^2}\right] \quad (2)$$

where $a$ controls the amplitude or strength of the input, $p$ controls the position of the input in the field, and $w$ controls the width of the input distribution (see Figure 5). $s_{target}$ represents the target voiceless distribution, and $s_{mp}$ represents the voiced minimal pair distribution. This treatment of speech intentions as *distributions* in feature space is similar to previous conceptualizations of speech production goals as "ranges" (Byrd & Saltzman, 2003), "windows" (Keating, 1990) or "convex regions" (Guenther, 1995). Each neuron $x'$ which exceeds an activation threshold contributes activation to other neurons $x$ via an interaction kernel $k(x - x')$ given by

$$k(x - x') = \frac{c_{exc}}{\sqrt{2\pi}\sigma_{exc}} \exp\left[-\frac{(x - x')^2}{2\sigma_{exc}^2}\right] - \frac{c_{inh}}{\sqrt{2\pi}\sigma_{inh}} \exp\left[-\frac{(x - x')^2}{2\sigma_{inh}^2}\right] - c_{glob} \quad (3)$$

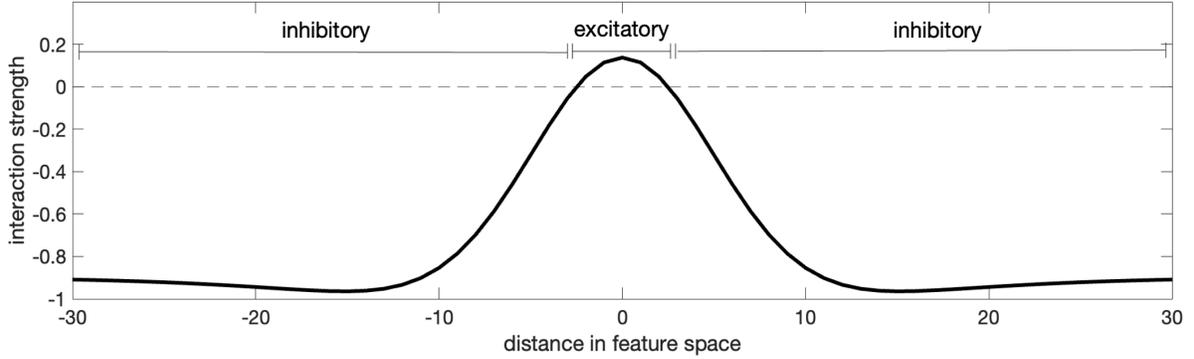

**Figure 3.** Lateral interaction kernel $k(x - x')$.

The effects of both excitatory and inhibitory interaction are modeled as Gaussian distributions centered on each neuron $x'$. $c_{exc}$ and $c_{inh}$ control the magnitude of excitatory and inhibitory interaction, respectively, and $\sigma_{exc}$ and $\sigma_{inh}$ control the width of each interaction distribution. $c_{glob}$ contributes additional across-the-board inhibition from each above-threshold neuron. In our model, $c_{exc} > c_{inh} > c_{glob}$ and $\sigma_{exc} < \sigma_{inh}$, so interaction is excitatory (positive effect on activation) for nearby neurons and inhibitory (negative effect on activation) for more distant neurons. Lateral excitation contributes to the stabilization of activation peaks which drive articulation, while lateral inhibition prevents runaway expansion of activation peaks. The activation threshold for interaction is given by a sigmoidal function $g(u)$, where $\beta$ controls the steepness of the threshold:

(4)

$$g(u) = \frac{1}{1 + \exp(-\beta u)}$$

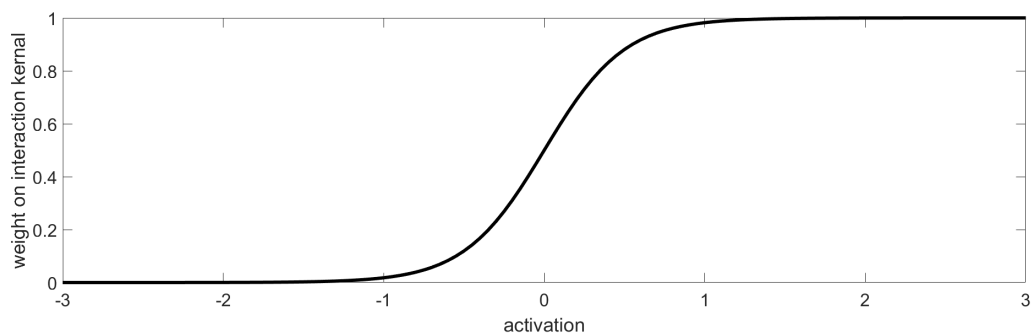

**Figure 4.** Sigmoidal function $g(u)$ gating lateral interaction; $\beta = 4$.

By convention, the threshold is set to $u = 0$ so that lateral interaction kicks in only when activation is positive. Finally, noise is simulated by adding normally distributed random values $\xi(x, t)$ weighted by a parameter $q$.

*2.3 Simulation results*

In this subsection, we use the model described above to simulate VOT planning for voiceless stop consonants in a number of conditions. The values of the field parameters used in all simulations are listed in Table 2, and the values of the input parameters are listed in Table 3. The input distributions are plotted in Figure 5.

| Parameter | Value |
|---|---|
| $\tau$ | 20 |
| $h$ | -5 |
| $\beta$ | 4 |
| $c_{exc}$ | 15 |
| $c_{inh}$ | 5 |
| $c_{glob}$ | 0.9 |
| $\sigma_{exc}$ | 5 |
| $\sigma_{inh}$ | 12.5 |
| $q$ | 1 |

Table 2. DNF parameter values.

| Input | Parameter | Value |
|---|---|---|
| $s_{target}$ (voiceless target) | $p$ | 70 |
| | $w$ | 30 |
| | $a$ | 6 |
| $s_{mp}$ (voiced minimal pair) | $p$ | 20 |
| | $w$ | 30 |
| | $a$ | varies |

Table 3. Input parameter values.

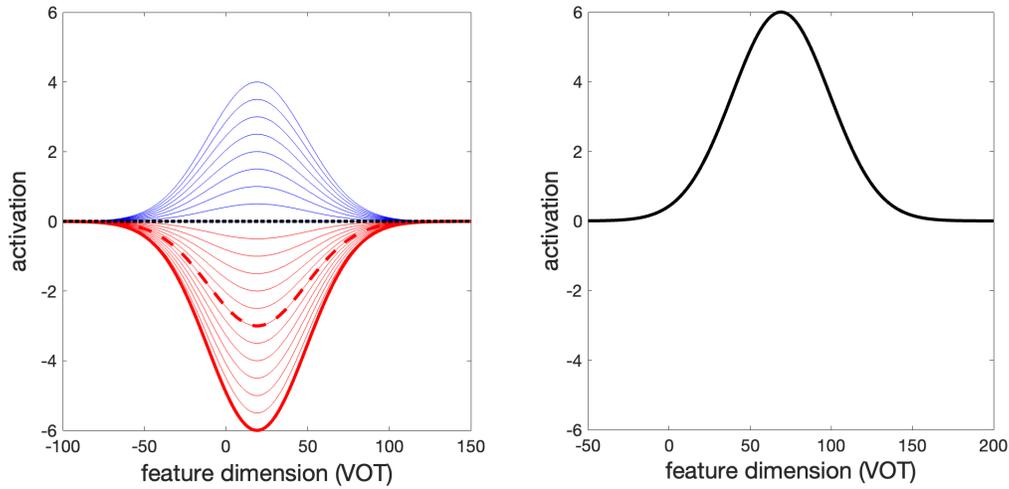

**Figure 5.** Right: voiceless target input distribution with constant positive amplitude $a$. Left: voiced minimal pair input distribution with varying $a$. Blue lines indicate positive values of $a$; red lines indicate negative values of $a$. The three bold lines correspond to the highlighted results in Figure 6.

Since part of our aim is to unify CH with trace effects in speech errors, we set the parameters of the DNF to be identical to those in Stern et al. (2022). This includes $\tau$, $h$, $\beta$, and $q$ as well as the parameters of the interaction kernel (see Figure 3). The global inhibition parameter $c_{glob}$ is large enough relative to the range of input amplitudes that we consider to ensure selection dynamics. That is, under these conditions, only a single peak will form regardless of the number of inputs to the field. This is crucial because human speech only allows the production of one VOT value per segment. The values of the input parameters (Table 3) were determined as follows. $p_{target}$ and $p_{mp}$ (the centers of the input distributions) were set to 70 ms and

20 ms, respectively, and both $w_{target}$ and $w_{mp}$ were set to 30, broadly consistent with recent reports of means and standard deviations measured for American English stop consonants (Chodroff & Wilson, 2017). $a_{target}$ was set to 6 in all simulations in order to ensure the formation of a stable activation peak in the context of a resting activation level $h = -5$. In order to investigate the effects of influence from a voiced minimal pair on VOT planning for a voiceless target, we varied $a_{mp}$ from –6 to 4 in steps of 0.5. At each value of $a_{mp}$, we simulated 500 instances of field evolution with 120 time steps each. The VOT target for each simulation was calculated as the point of maximum activation in the field at the final time step. Figure 6 displays the results.

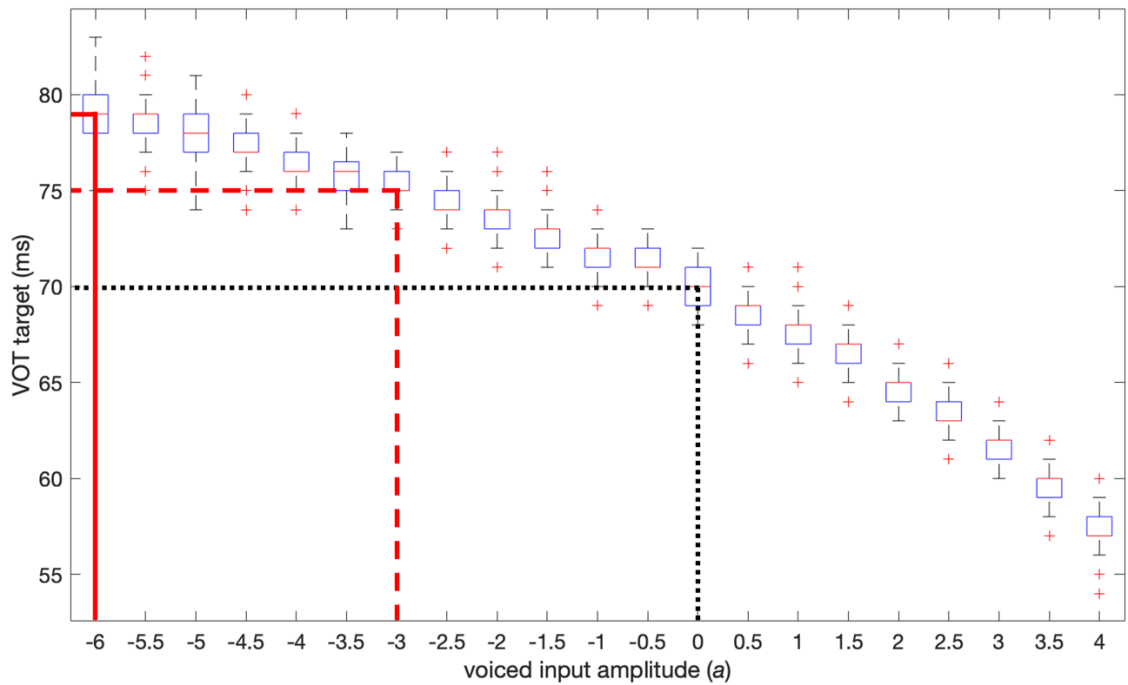

**Figure 6**. VOT targets generated by the DNF under the influence of two inputs: the voiceless target $s_{target}$ and the voiced minimal pair competitor $s_{mp}$ of varying amplitude, $a_{mp}$. Dotted black line: $a_{mp} = 0$, CH = 0 ("no competitor" condition from Baese-Berk & Goldrick, 2009). Dashed red line: $a_{mp} = -3$, CH ≈ 5 ("no context" condition). Solid red line: $a_{mp} = -6$, CH ≈ 10 ("context" condition).

Figure 6 demonstrates a clear trend whereby decreasing voiced input amplitude, $a_{mp}$ (increasing the magnitude of inhibitory input), corresponds to a larger (more hyperarticulated) voiceless VOT target. This is because $s_{target}$ and $s_{mp}$ partially overlap, so inhibitory input from $s_{mp}$ depresses activation in those regions of the field that correspond to a more *hypo*articulated voiceless VOT production. This is consistent with the empirical observation that CH of voiceless consonants is driven by a decrease in the likelihood of hypoarticulated productions, rather than an overall shift in the VOT distribution; i.e., CH is the result of a change in the skewness of the VOT distribution, rather than the mode (Buz et al., 2016). Increasing the amplitude of inhibitory input increases the magnitude of hyperarticulation: there is an approximately linear correlation between $a_{mp}$ and VOT target. The range of CH magnitudes observed in Figure 6 covers that observed in the three conditions from Baese-Berk & Goldrick (2009:Experiment 2), and allows us to understand the difference between these conditions as a difference in strength of inhibitory input from the minimal pair competitor. An example of field evolution in each of these conditions is displayed in Figure 7.

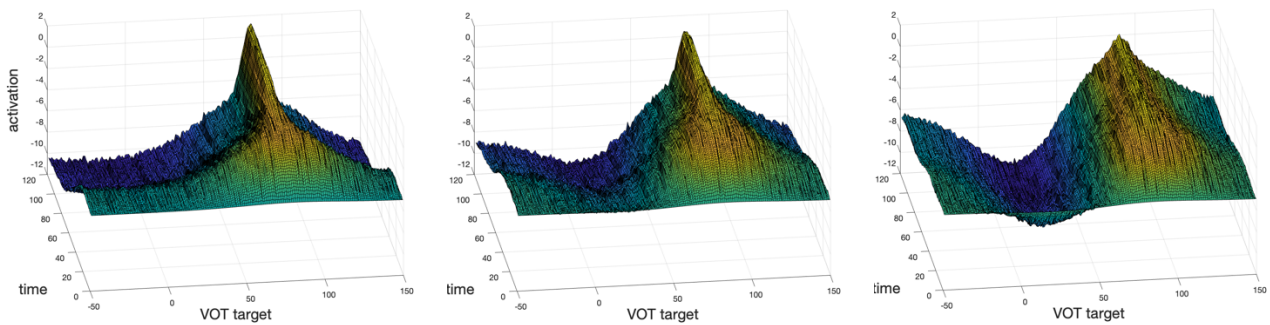

**Figure 7.** Examples of field evolution in the three conditions from Baese-Berk & Goldrick (2009:Experiment 2). Left: $a_{mp} = 0$, CH = 0 ("no competitor"). Middle: $a_{mp} = -3$, CH = 5 ("no context"). Right: $a_{mp} = -6$, CH = 10 ("context").

The field location of the activation peak shifts slightly depending on the magnitude of inhibitory voiced input: 70 ms (left), 75 ms (middle), and 80 ms (right). We can also observe that inhibitory voiced input *slows down* the formation of an activation peak. With no inhibitory input, the first neurons surpass the threshold for interaction ($u = 0$) at about the 40$^{th}$ time step, and the subsequent combination of local lateral excitation and global lateral inhibition causes an activation peak to form quite quickly. With intermediate inhibition, the first neurons do not surpass the threshold until about the 60th time step, and with strong inhibition, no neurons surpass the threshold until the very end of the 120-step simulation. In fact, in some simulations, 120 time steps was not enough time for any activation peak to form when $a_{mp} = -6$.[2] Thus, this model predicts a positive relationship between planning time and the magnitude of CH, a prediction which could be tested with an experiment designed to measure both response time and VOT.

Another observation from Figure 6 is that, when $a_{mp}$ is *positive*, we observe a *decrease* in VOT, i.e. a *hypo*articulation effect, consistent with the trace effect in speech errors (Stern et al., 2022). This framework thus allows us to understand both trace effects in speech errors and CH effects in non-errors as arising from the influence of lexical competitors on speech planning. The difference is that in the former case, this influence is excitatory, while in the latter case, it is inhibitory.

By understanding the range of previously observed CH effect magnitudes as arising from a cline of amplitudes of competitor inputs to a DNF governing pronunciation planning, we predict that intermediate input amplitudes should lead to intermediate CH effect magnitudes. For instance, VOT planning for a voiceless target under the influence of a voiced minimal pair competitor with input amplitude between 0 and –3 should lead to a *small* CH effect, i.e. smaller than 5 ms, the CH effect observed in the "no context" condition. One way to reduce the input amplitude of the competitor is to examine *pseudoword* production. Since pseudowords have no lexical representation, the network of interaction that activates the competitor has one less node, thereby decreasing the activation of the whole network, including the competitor (see Figure 2). In order to further vary the amplitude of competitor input, we can additionally introduce visual input to the competitor, similar to the "context" condition. The model predicts that CH should be observable for pseudowords, but it should be reduced in magnitude compared to CH in real words. We tested these predictions with an experiment, described in the following section.

---

[2] There are, in principle, different methods of interpreting the activation profile of the field in terms of an actionable target. Here, we selected the field position (neuron) with the highest activation level, which allows us to interpret field activation in terms of a VOT target even when no neuron breaches the interaction threshold. Some alternative methods, including integrating over above threshold neurons (Schöner et al., 2016) or selecting the first neuron to cross threshold (Harper, 2021), do not return a target value unless the field stabilizes.

## 3 Speech production experiment design

Our experimental design is largely a replication of Baese-Berk & Goldrick (2009:Experiment 2) with two main differences: target items were pseudowords, rather than real words, and the experiment was conducted over Zoom, rather than in the lab.

*3.1 Data availability*

All figures and analyses reported below were done in R version 4.1.2 (R Core Team, 2021). The anonymized data and R code are available on OSF at https://osf.io/hz8fp/.

*3.2 Participants*

24 adults participated in the experiment (ages 18-36, $M = 24.88$, $SD = 4.97$; 17 women, 7 men). This number of participants is twice that included in Baese-Berk & Goldrick (2009:Experiment 2), and was chosen in order to maximize the likelihood of detecting an effect (minimize the likelihood of a Type II error), given an expected increase in noise in online compared to in-person data collection. All participants self-reported that they were native speakers of American English, and that they had no history of speech, language, or hearing impairment. All participants provided informed consent under Yale University IRB #2000030436.

*3.3 Materials*

The experimental stimuli consisted of mono- and di-syllabic pseudowords beginning with voiceless stop consonants (see Appendix A for the complete list of stimuli). Two independent variables were manipulated: (1) whether or not the pseudoword had a minimal pair competitor in word-initial stop consonant voicing (e.g., TIVE /taɪv/, which has a minimal pair competitor DIVE /daɪv/ in the lexicon, vs. TIBE /taɪb/, which lacks a minimal pair competitor DIBE /daɪb/ in the lexicon) and (2) whether or not the minimal pair was salient in the context, i.e., presented as a competitor on the screen or not. Crossing these factors yields four experimental conditions. There were 12 target items per condition (four items for each major place of articulation: labial (/p/), alveolar (/t/), and velar (/k/)), for a total of 48 experimental target items. For the same reasons described above for participant selection, this number of experimental target items was chosen to be a substantial increase over the number (36) included in Baese-Berk & Goldrick (2009:Experiment 2). Each target item in the "minimal pair in lexicon" condition was matched for initial consonant and vowel with a target item in the "no minimal pair in lexicon" condition. Target items were balanced across levels of both independent variables for a number of phonotactic probability and phonological neighborhood measures (Vitevitch & Luce, 2004), summarized below:

- Sum segmental probability: log-transformed probability of each phoneme appearing in that position in the word, summed over all phonemes in the word

- Sum biphone probability: log-transformed probability of each sequence of two adjacent phonemes (biphone) appearing in that position in the word, summed over all biphones in the word

- Neighborhood density: count of real words that can be created by adding, removing, or changing one phoneme in the word

- Neighborhood frequency: mean frequency of occurrence of all phonological neighbors

The results of Welch's *t*-tests comparing each control measure between levels of the independent variables are displayed in Table 4.

**Table 4**. Results of Welch's *t*-tests comparing control variables between levels of the two independent variables. Measures are from (Vitevitch & Luce, 2004).

|  | Minimal pair in lexicon vs. no minimal pair in lexicon | | | Minimal pair on screen vs. no minimal pair on screen | | |
| --- | --- | --- | --- | --- | --- | --- |
|  | *t* | *df* | *p* | *t* | *df* | *p* |
| Sum segmental probability | –0.14 | 45.81 | 0.89 | 0.41 | 44.98 | 0.69 |
| Sum biphone probability | 0.62 | 41.68 | 0.54 | 0.71 | 44.09 | 0.48 |
| Neighborhood density | 0.02 | 45.28 | 0.98 | –0.49 | 43.47 | 0.63 |
| Neighborhood frequency | –0.71 | 33.76 | 0.49 | –1.45 | 23.35 | 0.16 |
| Sum segmental probability of minimal pair | –0.17 | 45.82 | 0.87 | 0.46 | 45.58 | 0.64 |
| Sum biphone probability of minimal pair | 0.61 | 41.46 | 0.54 | –0.32 | 41.02 | 0.75 |
| Neighborhood density of minimal pair | –0.78 | 44.05 | 0.44 | –1.08 | 45.87 | 0.29 |
| Neighborhood frequency of minimal pair | –1.35 | 37.60 | 0.19 | 0.49 | 39.77 | 0.63 |

24 filler items were also included, for a total of 72 items. Six filler items began with /s/ (e.g. SIP), six began with /ʃ/ (e.g. SHIP), and 12 began with a voiced stop consonant (e.g. DESK). Those beginning with voiced stop consonants were all minimal pair competitors of experimental stimuli, while the filler items beginning with /s/ and /ʃ/ were unrelated to the experimental items. Filler items included both real and pseudowords, and, like the experimental items, varied according to whether they had a minimal pair competitor in the lexicon or not and whether the minimal pair competitor was present on the screen or not.

*3.4 Procedure*

The experiment was conducted over a Zoom call hosted by the experimenter. Before the experiment, participants were instructed to join the Zoom call from a computer (not a phone) in a quiet room with a good internet connection. At the beginning of the experiment, the experimenter changed the participant's displayed name in the Zoom call to "speaker", and changed their own name to "experimenter". A lab confederate was also present in the call, whose Zoom name was changed to "listener". The participant (henceforth "speaker") was told that the listener was another naïve participant. Everyone in the Zoom call kept their cameras off but their microphones on for the duration of the experiment.

All instructions and stimuli were presented on a slideshow using screen sharing. During each trial, three words were presented on the shared screen: the target and two competitors, as seen in Figure 8. Before the experiment began, the speaker was sent a pdf that matched the shared screen, except that the target word for each trial was bolded and underlined. At the start of the experiment, the speaker was instructed to arrange their computer screen so that they could see both this pdf and the shared Zoom screen at the same time. Each time the experimenter advanced to the next trial, the listener cued the speaker by saying "ready". Then, the speaker produced the target word in the phrase "type the ___ number". This phrase was chosen in order to encourage fluent pronunciation. Words produced in isolation, i.e., not in a larger phrase, tend to have a slower "clear speech" pronunciation, which has previously been argued to obscure CH effects (Buz et al., 2016; Wedel et al., 2018). In this phrase, the target word does not occur at a major prosodic boundary, which could condition lengthening of the target word. Moreover, the speaker was instructed not to speak slowly or pause between words, but rather to speak at a quick, conversational pace. The listener then typed the number that corresponded to the word they heard ("1" = left, "2" = center, "3" = right) into the Zoom chat. If the response was correct, the experimenter played a bell sound; if it was incorrect, the experimenter played a buzz sound. The listener (a lab confederate) did not have independent access to the correct answers, and was simply instructed to participate naturally in the experiment. In order to increase motivation, participants were informed that they would receive an extra $5 if they finished the experiment with an

accuracy of 95% or higher (all participants finished above 95%, most at 100%). The complete instructions are included in Appendix B.

**A**

```
      1              2              3

   todge          shafe          dodge
```

**B**

```
      1              2              3

   todge          shafe          dodge
```

**Figure 8**. A: Example stimulus display from the "minimal pair in lexicon", "minimal pair on screen" condition. B: The corresponding display in the speaker's pdf.

After five practice trials with stimuli unrelated to the experimental stimuli, participants were instructed to ask the experimenter any questions they had. Then, the experimenter began recording the Zoom call, and each stimulus was presented twice, for a total of 144 trials (96 experimental, 48 fillers). The "minimal pair on screen" condition alternated between the two presentation lists, such that each stimulus was presented once with its minimal pair as an on-screen competitor, and once without. Four pseudo-randomizations of each presentation list were created, varying in the presentation order of the stimuli and in the relative positions of the target and competitors on the screen. Each of the six possible relative screen positions were used an equal number of times in each pseudo-randomization. In each pseudo-randomization, no two consecutive targets were minimal pairs (in any phonological feature in any position in the word) or had the same initial consonant and vowel; no two consecutive trials had the same two competitors; and no three consecutive trials had the same relative screen positions. Each pseudo-randomization of the first presentation list was combined with each pseudo-randomization of the second presentation list to create 16 unique combinations. Then, the presentation order of the two lists was varied to create 32 unique combinations. Each of the 24 participants thus completed the experiment with a unique stimulus presentation order. Participants were given a short break after each quarter of the experiment (three breaks total). The entire experiment lasted less than 30 minutes.

*3.5 Data processing*

Each participant in the Zoom call was individually recorded, allowing the creation of a wav file including only the speaker's audio. Silences between trials were removed from the wav file in Praat (Boersma & Weenink, 2021) using a custom script (Lennes, 2017), and the resulting file was force-aligned at the word and segment levels using the Montreal Forced Aligner (McAuliffe et al., 2017) with a customized English dictionary including phonetic transcriptions of the pseudoword stimulus items. Next, using AutoVOT (Sonderegger & Keshet, 2012), an automated VOT measurement algorithm was trained using 216 hand measurements in Praat from the onset of the release burst to the first zero-crossing of periodic vocal fold vibration: one measurement from each following vowel height (high, mid, low) from each consonant (/p/, /t/, /k/) from each subject (3 x 3 x 24). Even in ambiguous cases (e.g., when the release burst overlapped with some periodic energy from the preceding vowel, or when there was still some noise at the onset of periodicity), release burst and onset of periodic energy were always used as the indicators of the two events of interest. Since the AutoVOT algorithm requires the segment boundaries in the input TextGrids to be wider than the acoustic boundaries, we used a set of custom Praat scripts (Chodroff, 2019) to lengthen the segment boundaries of the word-initial voiceless stop consonants that were output by the Montreal Forced Aligner. Then, the trained algorithm was used to automatically measure the VOT of all 2,304 experimental tokens. In order to account for the effect of speech rate on VOT (Kessinger & Blumstein, 1997), a measure of speech rate was calculated as the duration in milliseconds from onset to offset of the trial (onset of "type" to offset of "number").

Trials were removed from analysis for the following reasons: (1) the speaker was disfluent or there was an interruption from background noise or internet connectivity issues (138 trials, 5.99%), (2) the listener responded inaccurately (6 trials, 0.26%), (3) VOT was less than 25 ms (likely a categorical error) (91 trials, 3.95%), or (4) speech rate was greater than three standard deviations from the mean (21 trials, 0.91%). In total, 2,048 trials (89.89%) were retained for analysis.

## 4 Experiment results

Before presenting the main results of the experiment, analyzed using Bayesian mixed effects regression models, we first present an analysis of the control variables that enter into the model.

*4.1 Principal component analysis of phonotactic probability and phonological neighborhood measures*

Although target items were balanced across the four experimental conditions for phonotactic probability and phonological neighborhood measures (as seen in Table 4), it is still useful to include these measures as control predictors in the mixed effects regression. However, as seen in Figure 9, there are a number of strong correlations between these control measures. These correlations could lead to a multicollinearity issue in the regression. In order to address this issue, we conducted principal component analysis (PCA) to identify the directions of maximal variability in the eight-dimensional space defined by these variables. The scree plot in Figure 10 reveals an elbow at component 3, suggesting retention of the first two components. These two components account for 64% of the variance in the original dataset.

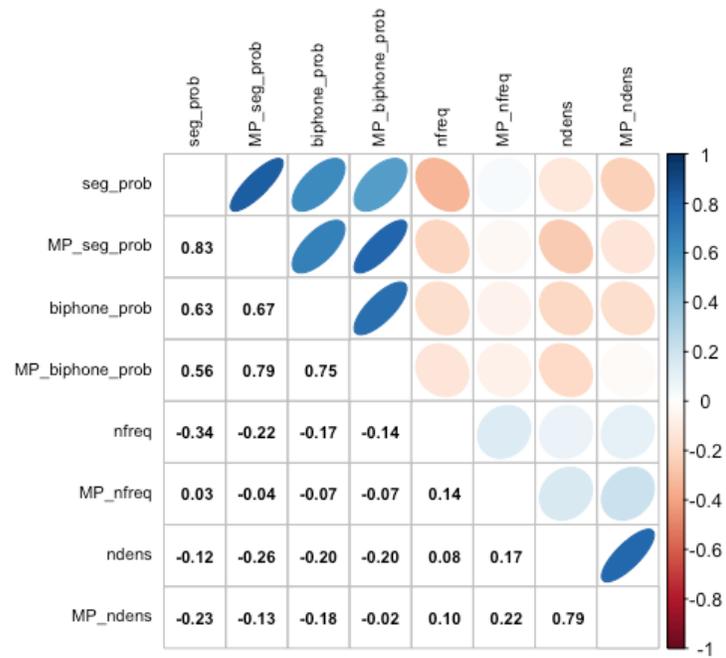

**Figure 9**. Correlations among lexical control measures.

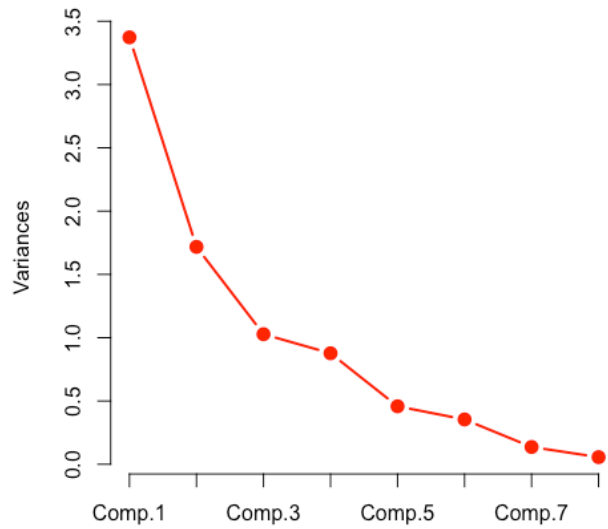

**Figure 10.** Scree plot of PCA results.

The loadings of the eight original variables on the two retained components—displayed in Table 5—suggest the following interpretation: component 1 generally indexes phonotactic probability (sum segmental probability and sum biphone probability of the target and minimal pair), while component 2 generally indexes phonological neighborhood density of the target and minimal pair. Importantly for the linear mixed effects regression, these two components are orthogonal to each other, addressing the issue of multicollinearity.

**Table 5.** Loadings of the eight original variables on the two retained components. The highest loadings on each component are bolded.

| Variable | Comp.1 | Comp.2 |
|---|---|---|
| sum segmental probability | **0.46** | 0.15 |
| sum biphone probability | **0.46** | 0.13 |
| neighborhood density | -0.23 | **0.61** |
| neighborhood frequency | -0.19 | 0.03 |
| sum segmental probability of minimal pair | **0.49** | 0.16 |
| sum biphone probability of minimal pair | **0.45** | 0.21 |
| neighborhood density of minimal pair | -0.2 | **0.65** |
| neighborhood frequency of minimal pair | -0.07 | 0.31 |

*4.2 Bayesian mixed effects model of VOT*

To analyze the effect of experimental condition on VOT, we performed Bayesian mixed effects regression using the *brms* package (Bürkner, 2017) in R (R Core Team, 2021). The model included the following control predictors: components 1 and 2 from the PCA, speech rate, trial number, and place of articulation. Place of articulation was treatment coded with "labial" as the reference level. The other measures were scaled and centered. The experimental fixed factors were minimal pair existence, minimal pair salience, and their interaction. Both experimental fixed factors were treatment coded with "false" as the reference level. The model included random slopes by subject for all experimental fixed factors, and a random slope by item for minimal pair salience.

We followed recommendations from Franke & Roettger (2019) for fitting Bayesian mixed models. We used the default priors of the *brms* package: a Student's *t*-distribution ($v = 3$, $\mu = 69$, $\sigma = 20.8$) for the intercept, a Student's *t*-distribution ($v = 3$, $\mu = 0$, $\sigma = 20.8$) for the standard deviation of the likelihood function and the random effects, and unbiased ("flat") priors for regression coefficients. We ran four sampling chains for 2000 iterations with a warm-up period of 1000 iterations for each model. All *R-hat* values (a diagnostic for convergence) were 1.0, indicating that the chains mixed successfully. Next, we removed observations with residuals to the model fit greater than 3 standard deviations from the mean (17 observations, 0.83%), and re-ran the model. Below we report the expected values of each regression coefficient under the posterior distribution and their 95% credible intervals (CrI). We consider it compelling evidence that a fixed factor significantly influenced VOT if the 95% CrI of the posterior distribution of the factor's coefficient does not overlap with 0. First, we address the effects of the control predictors. Component 1 (phonotactic probability) significantly decreased VOT, such that more phonotactically probable stimuli were produced with shorter VOT ($\beta = -1.16$, 95% CrI = [$-1.88$, $-0.42$]). The effect of component 2 (neighborhood density) did not reach significance ($\beta = 0.43$, 95% CrI = [$-0.61$, $1.51$]). Speech rate significantly increased VOT, such that VOT was longer at slower speech rates ($\beta = 7.07$, 95% CrI = [$6.34$, $7.81$]). Trial number significantly decreased VOT: as the experiment continued, participants tended to produce shorter VOT ($\beta = -1.20$, 95% CrI = [$-1.83$, $-0.57$]). Finally, place of articulation significantly affected VOT, such that alveolar (/t/: $\beta = 13.64$, 95% CrI = [$10.24$, $16.96$]) and velar (/k/: $\beta = 12.60$, 95% CrI = [$9.19$, $15.89$]) consonants had longer VOT than labial consonants (/p/). Turning to the experimental predictors, there was no significant main effect of either minimal pair existence ($\beta = 0.04$, 95% CrI = [$-3.25$, $3.39$]) or minimal pair salience ($\beta = -0.29$, 95% CrI = [$-2.10$, $1.49$]). However, there was a significant positive interaction: VOT was increased by a minimal pair competitor on the screen, but only when that minimal pair competitor was also a real word ($\beta = 2.50$, 95% CrI = [$0.04$, $4.93$]). As predicted, the magnitude of this CH effect was small compared with the previously observed effect on real words: 2.5 ms. The interaction is visualized in Figure 11.

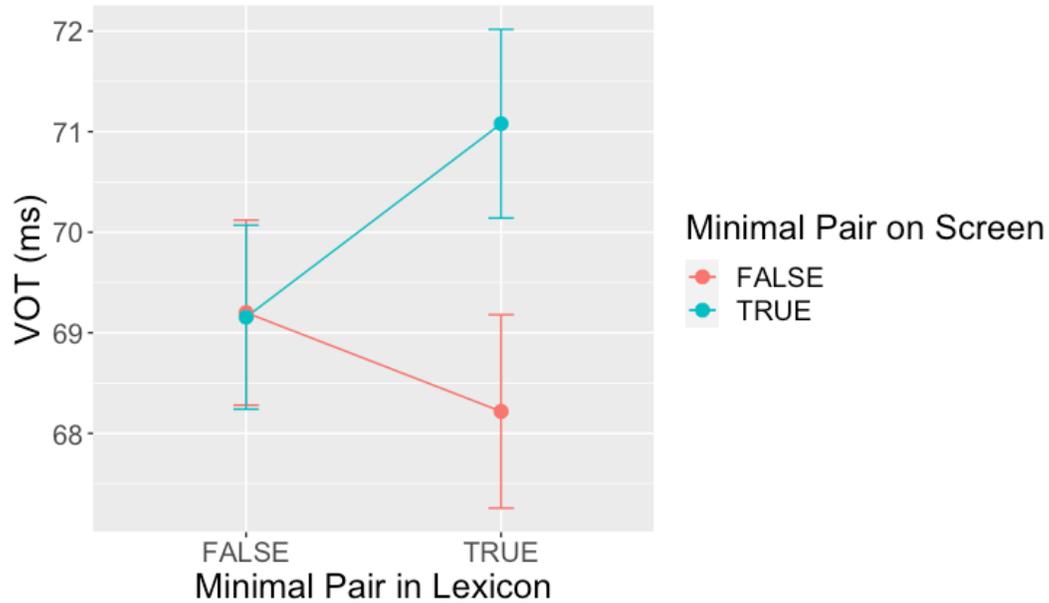

**Figure 11**. Mean VOT by condition. Error bars indicate standard error of the mean.

*4.3 Discussion*

The experiment results are consistent with a basis for CH in the real-time planning and production of speech, since pseudowords have no long-term lexical representations. Importantly, the CH effect in pseudowords was smaller in scope (only appearing when there was a real word minimal pair that was also present as an on-screen competitor) and magnitude (model estimate = 2.5 ms) compared to the CH effects observed with real words, i.e., 5 ms when the minimal pair was not contextually salient and 10 ms when it was (Baese-Berk & Goldrick, 2009). The reduced scope and magnitude of the pseudoword CH effect is consistent with reduced influence from the minimal pair competitor on pronunciation planning, due to reduced activation from a lexical level of planning.

## 5   General discussion

Cognitive mechanisms underlying speech production have been increasingly informed by the details of speech pronunciation. We proposed that a specific type of pronunciation variation, known as contrastive hyperarticulation (CH), is rooted in inhibitory neural connections between competitor lexical representations[3] and processes of pronunciation planning. A computational model of this proposal, developed within the framework of Dynamic Field Theory (DFT), derived the results of past studies and made new predictions. Specifically, varying competitor strength—i.e, the amplitude, $a$, of the competitor lexical item's input to a dynamic neural field (DNF) governing pronunciation planning—derived the effect size in our experiment as well as the conditions in Baese-Berk & Goldrick (2009:Experiment 2). With competitor input amplitude at –1.5, we derived our results. Increasing the amplitude of inhibitory input to –3 derived the effect found for real words in the "no context" condition of Baese-Berk & Goldrick (2009:Experiment 2). Increasing inhibitory input amplitude further to –6 derived the effect found for real words in the "context" condition of Baese-Berk & Goldrick (2009:Experiment 2). Thus, the variety of observed CH effect magnitudes can be derived by scaling a single model parameter, competitor strength. In addition, positive (excitatory) values of this parameter derive *hypo*articulation or "trace" effects of varying magnitudes, as observed in speech errors (Stern et al., 2022). Our theory thus offers a unified explanation of these apparently disparate phenomena. The model also makes predictions regarding response time, specifically: (i) a positive correlation between competitor strength and response time, and (ii) a positive correlation between response time and the magnitude of CH. These predictions remain to be tested. Broadly, the explicit incorporation of both temporal and feature gradience into a single model of pronunciation planning allows the generation of a rich and precise set of empirical predictions (Roon & Gafos, 2016). For instance, in the model, activation peaks narrow over time via the combination of lateral excitation and lateral inhibition. This predicts a relationship between response time and response variability, such that—all else equal—shorter response times should correspond with more variable responses, because the activation peak corresponding to the planned response has less time to narrow. Such a relationship between response time, width of neural activation peak, and response variability has been observed in manual reaching movements (Erlhagen & Schöner, 2002) and rhesus monkey motor mortex (Georgopoulos et al., 1986)—it remains to be tested in human speech.

---

[3] It is worth highlighting the fact that we model inhibition as coming from *lexical* representations, rather than sublexical representations like phonemes or syllables. The reason for this is empirical: CH (as well as other forms of context-based phonetic enhancement and reduction: Hall et al., 2018) appear to operate primarily on lexical items, rather than sublexical units. Both the /p/ in PET and the /p/ in PEP have a sublexical competitor /b/; CH is observed in the pronunciation of PET relative to PEP because of the presence of a *lexical* competitor in the former (BET) but not the latter (*BEP) case.

A reviewer raised the possibility that it is the relationship *between* activation of the target word and activation of the minimal pair competitor—rather than activation of the minimal pair competitor, per se—that determines the pronunciation outcome. In order to investigate this issue, we ran simulations using the same parameters described in Section 2, but varying both competitor *and* target input amplitude. We varied the competitor input amplitude, $a_{mp}$, from –6 to 5 and the target input amplitude, $a_{target}$, from 5 to 10, both in steps of 0.5. We simulated 500 productions in each condition and recorded each VOT target. Figure 12 displays the results. Consistent with the results in Section 2.3, the primary predictor of the qualitative behavior of the system is the competitor input amplitude, $a_{mp}$. When $a_{mp}$ is negative, CH is observed; when $a_{mp}$ is positive, a trace effect is observed. The magnitude of each type of effect correlates with the absolute value of $a_{mp}$. In contrast, the target input amplitude, $a_{target}$, modulates the magnitude of competitor influence. Greater $a_{target}$ corresponds with less competitor influence, i.e., smaller CH when competitor input is inhibitory, and a smaller trace effect when competitor input is excitatory. Interestingly, there is an asymmetry between CH and trace effects in this regard. The magnitude of CH is relatively robust to changes in $a_{target}$: when competitor input is strongly inhibitory ($a_{mp}$ = –6)[4], mean CH ranges from 6.1 ms ($a_{target}$ = 10) to 10.4 ms ($a_{target}$ = 5). However, when competitor input is strongly excitatory ($a_{mp}$ = 5), the magnitude of the trace effect ranges from 8.4 ms ($a_{target} = 10$) to 24.9 ms ($a_{target} = 5$). This is because as $a_{mp}$ approaches $a_{target}$, the two hills of activation begin to merge into a single peak, approximately equidistant from the centers of the two input distributions $p_{mp}$ and $p_{target}$. No such mechanism operates when $a_{mp}$ is negative. Thus, while these post-hoc simulations support our claim that the magnitude of CH is primarily determined by activation of a minimal pair competitor, it also suggests the need for follow-up work investigating the influences of target and competitor activation in relation to CH and trace effects.

---

[4] VOT targets reported in the simulation include some cases in which the field did not stabilize within the timeframe of the simulation, which is possible given our method of interpreting field dynamics in terms of VOT targets (see footnote 3).

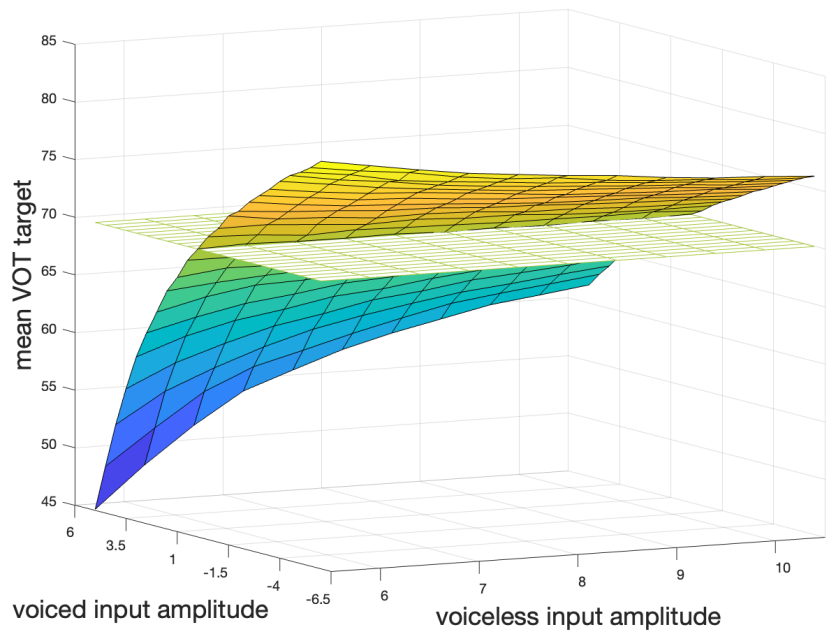

**Figure 12.** Mean VOT target (z-axis) by voiceless target input amplitude $a_{target}$ (x-axis) and voiced competitor input amplitude $a_{mp}$ (y-axis). The center of the target distribution $p_{target} = 70$ ms is shown with a flat plane; values above this plane (yellow) indicate CH, and values below this plane (blue) indicate a trace effect.

In the model presented here, CH is derived from a mechanism internal to the speech production system, i.e., inhibitory input from competitor lexical representations to pronunciation planning fields. The listener has no explicit role in the model. However, the speaker-internal inhibitory mechanism is compatible with audience design theories of CH (Buz et al., 2016; Munson & Solomon, 2004; Schertz, 2013; Wright, 2004). Inhibition can be seen to *implement* audience design on the part of the speaker, since inhibition has the effect of making pronunciations more easily perceivable for the listener. In this way, a potential theoretical dichotomy between "speaker-internal" vs. "audience design" mechanisms of CH can be resolved. The mechanism itself is speaker-internal, but it may serve a listener-oriented function. This conceptualization has the potential to shed light on previously observed listener-driven modulation of CH effects. For instance, Buz et al. (2016) observed an increase in the magnitude of CH following explicit misrecognitions by a simulated listener. In the present framework, there are at least two potential mechanisms that might drive this effect. First, it is possible that as speakers sense an increased risk of listener misrecognition, this causes an increase in the activation of competitors, increasing the magnitude of inhibitory input from competitors to pronunciation planning fields, thus increasing the magnitude of CH. This mechanism would capture the intuition that strongly intending *not* to say a word actually *increases* mental activation of that word, and is consistent with the observation in Buz et al. (2016) that listener-driven modulation of CH was only observed on trials where the minimal pair competitor was present. Another possibility is that increased speaker awareness of the risk of listener misrecognition modulates the function relating lexical activation to field input, leading to greater inhibitory input given the same magnitude of competitor activation (see further discussion of this function below). Further modeling and empirical work would be necessary to refine and distinguish between these proposed mechanisms.

The difference in CH magnitude between real words and pseudowords supports a role for lexical representations in contributing to CH. So far, we have considered the hypothesis that this contribution takes place primarily during the planning of individual utterances, increasing the strength of inhibitory input from competitor lexical items to pronunciation planning fields. However, another possibility, discussed in Section 1.2, is that lexical representations contribute to CH on a slower timescale, via updating of detailed episodic memories of perceived and produced pronunciations (e.g., Goldinger, 1998; Pierrehumbert, 2001, 2002). In the present framework, this amounts to lexical representations updating their input *distributions*, parameterized as $p$ and $w$. It is possible that the coupling relationships between lexical representations and pronunciation planning fields are somewhat idiosyncratic, such that each lexical representation projects a unique set of inputs to pronunciation planning fields. These coupling relationships might change subtly over the lifetimes of language users based on experience (Gafos & Kirov, 2009), leading to the observed idiosyncratic phonetic differences between lexical items (e.g., Tang & Shaw, 2021). It is possible that lexical representations contribute to CH via mechanisms on both timescales: interactive activation during speech planning, and long-term shifts in distributions of episodic memories. A combination of real-time and long-term effects on pronunciation details has previously been observed in the relationship between contextual predictability and phonetic cues of prominence like duration, pitch, and intensity (Seyfarth, 2014; Sóskuthy & Hay, 2017; Tang & Shaw, 2021). The contribution of a long-term mechanism is possibly apparent in the fact that VOT was overall slightly higher in the real words from Baese-Berk & Goldrick (2009) (baseline condition = 72 ms) compared to the pseudowords from the present experiment (baseline = 69 ms). However, the present experiment was not an exact replication of Baese-Berk & Goldrick (2009); most notably, our experiment was conducted over Zoom rather than in person. This may have affected the degree of clear speech employed by participants, which has been argued to affect the magnitude of CH effects (Wedel et al., 2018). Thus, it is difficult to interpret any direct quantitative comparison between the results of our study and those of Baese-Berk & Goldrick (2009).

Deriving the range of observed CH and trace effects from competitor strength raises a number of novel research questions. In our simulations, we varied competitor strength systematically to observe its effects on pronunciation. Presumably, the ranges of competitor strength derive from other factors that can be built into a more elaborate model. In future work, we plan to derive competitor strength and polarity directly from the lexicon. This will enable us to define the magnitude of *a* as a function of lexical activation, such that more active lexical items project stronger input to pronunciation planning DNFs. We also posited that the *polarity* of *a* (i.e., excitatory vs. inhibitory) varies, deriving the difference between CH (from inhibitory competitor input) and trace effects (from excitatory competitor input). What determines the polarity of *a*? One possibility is that input polarity is a function of the similarity of the *current* state of the field to the lexical representation's *preferred* state of the field, i.e., the state that *would* be induced by excitatory input from that lexical representation. For instance, in the example in Figure 2, since PUN becomes active before BUN, by the time BUN is active enough to affect pronunciation planning DNFs, the VOT field is in a state more consistent with a voiceless production than a voiced production. This induces the BUN-to-VOT projection to be inhibitory. These elaborations would derive competitor strength and polarity directly from the lexicon, leading to new predictions for how pronunciation varies across words.

Another future direction involves deriving speech errors. Noise in the relationship between field state and input polarity offers a possible mechanism underlying speech errors: sometimes, lexical-to-phonological projection is excitatory, even when the field state differs from the lexical item's preferred state, leading to a trace effect. The fact that visual input to the minimal pair competitor increases the magnitude of CH (by hypothesis, by increasing the magnitude of inhibitory lexical-to-phonological input) supports the notion that the *magnitude* of input is a positive monotonic function of lexical activation, regardless of the *polarity* of input. Noise around the polarity of the input could cause a strong inhibitory input to flip to a strong excitatory input, driving the field towards a state consistent with the pronunciation of a lexical competitor, i.e., a speech error.

In our account, CH derives from decreased activation in certain regions of a pronunciation planning field caused by inhibitory input from a minimal pair competitor. A reviewer raises the possibility that this metrically specific decrease in neural activation might be caused by a different mechanism, like habituation via synaptic fatigue. Pursuing this possibility would raise a number of interesting questions. For instance, why would competitor activation cause synaptic fatigue in fields governing dimensions of pronunciation on which the competitor differs from the target word, but not in fields governing dimensions on which the competitor and target overlap? Moreover, how do patterns of synaptic fatigue differ between non-errorful and errorful speech, such that CH occurs in the former but trace effects occur in the latter? We feel that, at this point, our inhibition account is the most promising in terms of coverage of existing empirical facts and generation of novel predictions, but it would be useful to carefully formulate alternatives in order to compare their predictions with those of our account.

To take a broader theoretical perspective on our proposal, we can ask why the neurocognitive mechanisms underlying speech planning would have evolved to incorporate both excitatory and inhibitory interactions between lexical and phonological levels of planning. In other words, why would activating one lexical representation affect the activation of other similar lexical representations (competitors), and why would competitors affect pronunciation planning? From the perspective of production, interactive lexical-phonological activation can be seen to have a facilitative effect, speeding up the process of activation peak stabilization in pronunciation planning fields via additional input from competitors. Thus, having competitors facilitates pronunciation. Of course, activating lexical competitors and allowing them to influence pronunciation planning introduces a risk that the planning process might converge on a competitor pronunciation, rather than the target. In fact, the cost of excitatory interaction is likely often seen in speech errors, when a competitor is produced instead of the target. Excitatory projections bring the benefit of faster speech production and the risk of occasional errors. Inhibitory lexical-to-phonological projection can be seen to mitigate this risk. By projecting inhibitory input to just those pronunciation planning fields that differentiate the competitor from the target (via the mechanism described above), competitors are able to facilitate planning of dimensions of pronunciation which are shared with the target without disrupting planning of dimensions which are not shared with target. Moreover, as discussed above, this inhibitory mechanism serves the additional function of making pronunciations more easily perceivable by listeners. Both excitatory and inhibitory coupling mechanisms between the lexicon and pronunciation planning fields have clear roles in a speech production system evolved for efficiency.

Explicitly incorporating these hypothesized lexical-phonological coupling mechanisms into the present model has the potential to shed new light on the mechanisms underlying the effects of phonological neighborhoods on speech production. Different studies have identified facilitation (Vitevitch, 2002), inhibition (Gordon & Kurczek, 2014), hyperarticulation (Munson & Solomon, 2004), and hypoarticulation (Gahl et al., 2012) induced by phonological neighborhood density. Perhaps this diversity of findings is a result of the coarseness of existing phonological neighborhood density measures: most commonly, the number of real words that can be created by adding, removing, or changing a single phoneme. In the framework presented here, neighborhood density itself is not expected to have a consistent effect on pronunciation planning. Rather, different types of neighbors are expected to have different effects on different dimensions of pronunciation. In particular, neighbors are predicted to inhibit dimensions that differentiate them from the target and excite dimensions on which they overlap with the target. The aggregate of these effects from all neighbors will ultimately derive the pronunciation targets of a particular utterance, and the time it takes to plan these targets. Crucially, the simple *number* of neighbors should be a weak predictor of target values and target planning times. For instance, a word with many more neighbors that *differ* in initial consonant voicing than neighbors that *share* initial consonant voicing is likely to have a hyperarticulated VOT, compared to a word with more neighbors that share voicing than that differ in voicing—even if both words have the same neighborhood density. The complexity of the influences of phonological neighborhoods on speech production points to the utility of a computational modeling approach like the one pursued here. A mathematical formalization of the processes of pronunciation planning and their coupling to lexical planning allows the generation of precise quantitative predictions in cases where such predictions could not possibly be intuited based on a verbally articulated theory alone. We believe this is a promising direction for future work and may lead to new empirical predictions linking the temporal dynamics of speech planning to the phonetic details of pronunciation.

# 6 Conclusion

We investigated the real-time mechanisms contributing to contrastive hyperarticulation (CH) during the planning of individual utterances. We demonstrated—using a model of pronunciation planning based on Dynamic Field Theory (DFT)—that CH is derivable from inhibitory projections from the minimal pair lexical representation to phonological levels of planning. We also showed that the magnitude of inhibitory projection correlates with the magnitude of CH, and that the magnitude of *excitatory* projection correlates with the magnitude of the phonetic trace effect in speech errors. We thus derived the observed range of CH and trace effect magnitudes by scaling a single model parameter, competitor strength, offering a unified explanation of an apparently disparate set of phenomena. We tested some predictions of the model with a novel experiment, very similar to Experiment 2 in Baese-Berk & Goldrick (2009) but with pseudoword stimuli. Pseudowords demonstrated CH effects, consistent with a contribution of real-time mechanisms to CH. However, the scope and magnitude of CH in pseudowords was reduced compared to CH in real words, consistent with a role for interactive activation between lexical and phonological representations. We outlined directions for future work which are suggested by the present study.

**Appendix A.** Stimuli.

**Table A1.** Experimental stimuli from presentation list 1. In list 2, the "minimal pair on screen" and "no minimal pair on screen" conditions are switched. Minimal pairs differing in word-initial consonant voicing are included in parentheses, but were not presented as target items except during filler trials.

|  | Minimal pair in lexicon | | | No minimal pair in lexicon | | |
|---|---|---|---|---|---|---|
|  | Labial | Alveolar | Velar | Labial | Alveolar | Velar |
| **Minimal pair on screen (list 1)** | 1. peam (beam) | 1. teth (death) | 1. keese (geese) | 1. peeb (beeb) | 1. tep (dep) | 1. keet (geet) |
|  | 2. pelch (belch) | 2. todge (dodge) | 2. cosh (gosh) | 2. peft (beft) | 2. tob (dob) | 2. codge (godge) |
|  | 3. pid (bid) | 3. tesk (desk) | 4. kig (gig) | 3. pim (bim) | 3. teld (deld) | 3. kip (gip) |
|  | 4. potch (botch) | 4. tive (dive) | 4. kulp (gulp) | 4. podge (bodge) | 4. tibe (dibe) | 4. kulk (gulk) |
| **No minimal pair on screen (list 1)** | Labial | Alveolar | Velar | Labial | Alveolar | Velar |
|  | 1. pag (bag) | 1. teff (deaf) | 1. coof (goof) | 1. paz (baz) | 1. teg (deg) | 1. koom (goom) |
|  | 2. pabble (babble) | 2. tid (did) | 2. kiv (give) | 2. packle (backle) | 4. tiv (div) | 2. kidge (gidge) |
|  | 3. pathe (bathe) | 3. tupe (dupe) | 3. kide (guide) | 3. pame (bame) | 3. toog (doog) | 3. kife (gife) |
|  | 4. pottle (bottle) | 4. tope (dope) | 4. kest (guest) | 4. possle (bossle) | 4. tobe (dobe) | 4. keft (geft) |

**Table A2.** Filler stimuli beginning with sibilants. Stimuli beginning with both /s/ and /ʃ/ were presented.

| Minimal pair in lexicon | No minimal pair in lexicon |
|---|---|
| 1. sack (shack) | 1. sap (shap) |
| 2. sip (ship) | 2. sick (shick) |
| 3. same (shame) | 3. safe (shafe) |
| 4. sin (shin) | 4. sing (shing) |
| 5. save (shave) | 5. saint (shaint) |
| 6. sore (shore) | 6. soap (shoap) |

**Table A3.** Unrandomized list of trials including target items and both competitors for practice trials and both presentation lists.

| list | trial type | minimal pair in lexicon | minimal pair on screen | place of articulation | target | competitor 1 | competitor 2 |
|---|---|---|---|---|---|---|---|
| practice | practice | — | — | — | vent | vine | chair |
| practice | practice | — | — | — | teef | teeth | phone |
| practice | practice | — | — | — | pite | pote | thick |
| practice | practice | — | — | — | beef | beeth | door |
| practice | practice | — | — | — | leel | real | plant |
| 1 | experimental | yes | yes | labial | peam | beam | sack |
| 1 | experimental | yes | yes | labial | pelch | belch | sip |
| 1 | experimental | yes | yes | labial | pid | bid | same |
| 1 | experimental | yes | yes | labial | potch | botch | shap |
| 1 | experimental | yes | yes | alveolar | teth | death | shick |
| 1 | experimental | yes | yes | alveolar | todge | dodge | shafe |
| 1 | experimental | yes | yes | alveolar | tesk | desk | sin |
| 1 | experimental | yes | yes | alveolar | tive | dive | save |
| 1 | experimental | yes | yes | velar | keese | geese | sore |
| 1 | experimental | yes | yes | velar | cosh | gosh | sing |
| 1 | experimental | yes | yes | velar | kig | gig | saint |
| 1 | experimental | yes | yes | velar | kulp | gulp | soap |
| 1 | experimental | yes | no | labial | pag | sack | shack |
| 1 | experimental | yes | no | labial | pabble | sip | ship |
| 1 | experimental | yes | no | labial | pathe | same | shame |
| 1 | experimental | yes | no | labial | pottle | sin | shin |
| 1 | experimental | yes | no | alveolar | teff | save | shave |
| 1 | experimental | yes | no | alveolar | tid | sore | shore |
| 1 | experimental | yes | no | alveolar | tupe | sap | shap |
| 1 | experimental | yes | no | alveolar | tope | sick | shick |
| 1 | experimental | yes | no | velar | coof | safe | shafe |
| 1 | experimental | yes | no | velar | kiv | sing | shing |
| 1 | experimental | yes | no | velar | kide | saint | shaint |
| 1 | experimental | yes | no | velar | kest | soap | shoap |
| 1 | experimental | no | yes | labial | peeb | beeb | shack |
| 1 | experimental | no | yes | labial | peft | beft | ship |
| 1 | experimental | no | yes | labial | pim | bim | shame |
| 1 | experimental | no | yes | labial | podge | bodge | sap |
| 1 | experimental | no | yes | alveolar | tep | dep | sick |

| 1 | experimental | no | yes | alveolar | tob | dob | safe |
|---|---|---|---|---|---|---|---|
| 1 | experimental | no | yes | alveolar | teld | deld | shin |
| 1 | experimental | no | yes | alveolar | tibe | dibe | shave |
| 1 | experimental | no | yes | velar | keet | geet | shore |
| 1 | experimental | no | yes | velar | codge | godge | sing |
| 1 | experimental | no | yes | velar | kip | gip | saint |
| 1 | experimental | no | yes | velar | kulk | gulk | soap |
| 1 | experimental | no | no | labial | paz | sack | shack |
| 1 | experimental | no | no | labial | packle | sip | ship |
| 1 | experimental | no | no | labial | pame | same | shame |
| 1 | experimental | no | no | labial | possle | sin | shin |
| 1 | experimental | no | no | alveolar | teg | save | shave |
| 1 | experimental | no | no | alveolar | tiv | sore | shore |
| 1 | experimental | no | no | alveolar | toog | sap | shap |
| 1 | experimental | no | no | alveolar | tobe | sick | shick |
| 1 | experimental | no | no | velar | koom | safe | shafe |
| 1 | experimental | no | no | velar | kidge | sing | shing |
| 1 | experimental | no | no | velar | kife | saint | shaint |
| 1 | experimental | no | no | velar | keft | soap | shoap |
| 1 | filler | yes | yes | ʃ | shack | sack | pag |
| 1 | filler | yes | yes | ʃ | ship | sip | pabble |
| 1 | filler | yes | yes | ʃ | shame | same | pathe |
| 1 | filler | yes | no | ʃ | shin | tesk | desk |
| 1 | filler | yes | no | ʃ | shave | tive | dive |
| 1 | filler | yes | no | ʃ | shore | keese | geese |
| 1 | filler | no | yes | s | sap | shap | tupe |
| 1 | filler | no | yes | s | sick | shick | tope |
| 1 | filler | no | yes | s | safe | shafe | coof |
| 1 | filler | no | no | s | sing | codge | godge |
| 1 | filler | no | no | s | saint | kip | gip |
| 1 | filler | no | no | s | soap | kulp | gulp |
| 1 | filler | yes | yes | labial | beam | peam | sack |
| 1 | filler | yes | yes | alveolar | dodge | todge | shafe |
| 1 | filler | yes | yes | velar | gig | kig | saint |
| 1 | filler | yes | no | labial | belch | same | shame |
| 1 | filler | yes | no | alveolar | desk | sore | shore |
| 1 | filler | yes | no | velar | geese | saint | shaint |
| 1 | filler | no | yes | labial | beeb | peeb | shack |
| 1 | filler | no | yes | alveolar | deld | teld | shin |

| 1 | filler | no | yes | velar | gosh | cosh | sing |
|---|---|---|---|---|---|---|---|
| 1 | filler | no | no | labial | bodge | sip | ship |
| 1 | filler | no | no | alveolar | dibe | sap | shap |
| 1 | filler | no | no | velar | geet | sing | shing |
| 2 | experimental | yes | no | labial | peam | sack | shack |
| 2 | experimental | yes | no | labial | pelch | sip | ship |
| 2 | experimental | yes | no | labial | pid | same | shame |
| 2 | experimental | yes | no | labial | potch | sin | shin |
| 2 | experimental | yes | no | alveolar | teth | save | shave |
| 2 | experimental | yes | no | alveolar | todge | sore | shore |
| 2 | experimental | yes | no | alveolar | tesk | sap | shap |
| 2 | experimental | yes | no | alveolar | tive | sick | shick |
| 2 | experimental | yes | no | velar | keese | safe | shafe |
| 2 | experimental | yes | no | velar | cosh | sing | shing |
| 2 | experimental | yes | no | velar | kig | saint | shaint |
| 2 | experimental | yes | no | velar | kulp | soap | shoap |
| 2 | experimental | yes | yes | labial | pag | bag | sack |
| 2 | experimental | yes | yes | labial | pabble | babble | sip |
| 2 | experimental | yes | yes | labial | pathe | bathe | same |
| 2 | experimental | yes | yes | labial | pottle | bottle | shap |
| 2 | experimental | yes | yes | alveolar | teff | deaf | shick |
| 2 | experimental | yes | yes | alveolar | tid | did | shafe |
| 2 | experimental | yes | yes | alveolar | tupe | dupe | sin |
| 2 | experimental | yes | yes | alveolar | tope | dope | save |
| 2 | experimental | yes | yes | velar | coof | goof | sore |
| 2 | experimental | yes | yes | velar | kiv | give | sing |
| 2 | experimental | yes | yes | velar | kide | guide | saint |
| 2 | experimental | yes | yes | velar | kest | guest | soap |
| 2 | experimental | no | no | labial | peeb | sack | shack |
| 2 | experimental | no | no | labial | peft | sip | ship |
| 2 | experimental | no | no | labial | pim | same | shame |
| 2 | experimental | no | no | labial | podge | sin | shin |
| 2 | experimental | no | no | alveolar | tep | save | shave |
| 2 | experimental | no | no | alveolar | tob | sore | shore |
| 2 | experimental | no | no | alveolar | teld | sap | shap |
| 2 | experimental | no | no | alveolar | tibe | sick | shick |
| 2 | experimental | no | no | velar | keet | safe | shafe |
| 2 | experimental | no | no | velar | codge | sing | shing |
| 2 | experimental | no | no | velar | kip | saint | shaint |

| 2 | experimental | no | no | velar | kulk | soap | shoap |
|---|---|---|---|---|---|---|---|
| 2 | experimental | no | yes | labial | paz | baz | shack |
| 2 | experimental | no | yes | labial | packle | backle | ship |
| 2 | experimental | no | yes | labial | pame | bame | shame |
| 2 | experimental | no | yes | labial | possle | bossle | sap |
| 2 | experimental | no | yes | alveolar | teg | deg | sick |
| 2 | experimental | no | yes | alveolar | tiv | div | safe |
| 2 | experimental | no | yes | alveolar | toog | doog | shin |
| 2 | experimental | no | yes | alveolar | tobe | dobe | shave |
| 2 | experimental | no | yes | velar | koom | goom | shore |
| 2 | experimental | no | yes | velar | kidge | gidge | sing |
| 2 | experimental | no | yes | velar | kife | gife | saint |
| 2 | experimental | no | yes | velar | keft | geft | soap |
| 2 | filler | yes | no | ʃ | shack | pag | bag |
| 2 | filler | yes | no | ʃ | ship | pabble | babble |
| 2 | filler | yes | no | ʃ | shame | pathe | bathe |
| 2 | filler | yes | yes | ʃ | shap | sap | tesk |
| 2 | filler | yes | yes | ʃ | shick | sick | tive |
| 2 | filler | yes | yes | ʃ | shafe | safe | keese |
| 2 | filler | no | no | s | sin | tupe | dupe |
| 2 | filler | no | no | s | save | tope | dope |
| 2 | filler | no | no | s | sore | coof | goof |
| 2 | filler | no | yes | s | sing | shing | codge |
| 2 | filler | no | yes | s | saint | shaint | kip |
| 2 | filler | no | yes | s | soap | shoap | kulp |
| 2 | filler | yes | no | labial | beam | sack | shack |
| 2 | filler | yes | no | alveolar | dodge | safe | shafe |
| 2 | filler | yes | no | velar | gig | saint | shaint |
| 2 | filler | yes | yes | labial | bottle | pottle | shap |
| 2 | filler | yes | yes | alveolar | did | tid | shafe |
| 2 | filler | yes | yes | velar | give | kiv | sing |
| 2 | filler | no | no | labial | beeb | sack | shack |
| 2 | filler | no | no | alveolar | deld | sin | shin |
| 2 | filler | no | no | velar | gosh | sing | shing |
| 2 | filler | no | yes | labial | baz | paz | shack |
| 2 | filler | no | yes | alveolar | deg | teg | sick |
| 2 | filler | no | yes | velar | geft | keft | soap |

**Appendix B.** Instructions.

Welcome to the experiment! Please leave your camera off but your microphone on (unmuted). The experimenter will randomly change one of your Zoom names to "speaker", and the other to "listener". "Speaker": you will receive a pdf in the Zoom chat. Please open this pdf to the first page, and arrange your computer screen so that you can see both the pdf and the shared Zoom screen at the same time.

Your goal is to communicate successfully by working together. (Please say "ok" when you've finished reading the slide).

Each trial in the experiment will go like this: Three words will appear on the screen. Some of the words will be real words of English, and others will be made up words.

Listener: After the words appear on the screen, say "ready". Then, the speaker will say the target word for you. If you think the target is the word on the left side of the screen, type "1" into the Zoom chat. If you think it is the word in the center, type "2". If you think it is the word on the right, type "3".

Listener: It is your task to listen closely to make sure you select the correct target word. During the experiment, please do not say anything except "ready".

Speaker: In the pdf you received, the target word for each trial is bolded and underlined. Before each trial, make sure the three words on the pdf match the three words on the shared Zoom screen.

Speaker: After the listener says "ready", say the target word in the phrase "type the ___ number", for example "type the sample number". If the target is not a real word of English, just pronounce it in the way that seems most natural. It is your task to make sure the listener chooses the correct word.

Speaker: Please do not speak slowly or pause between words. Rather, speak at a fluent, conversational pace. During the experiment, please do not say anything except the target word in the phrase "type the ___ number".

After each trial, you will hear a bell sound if you are correct, or a buzz sound if you are incorrect. If you finish the experiment with an overall accuracy of 95% or more, you will each receive a bonus $5. Let's do a little practice before we start.